\PassOptionsToPackage{table}{xcolor}

\documentclass{article} % For LaTeX2e
\usepackage{iclr2027_conference,times}

\usepackage{amsmath,amsfonts,bm}

\def\eqref#1{equation~\ref{#1}}
\def\1{\bm{1}}

\DeclareMathAlphabet{\mathsfit}{\encodingdefault}{\sfdefault}{m}{sl}
\SetMathAlphabet{\mathsfit}{bold}{\encodingdefault}{\sfdefault}{bx}{n}

\usepackage[table]{xcolor}

\usepackage{xcolor}
\usepackage{wrapfig}
\usepackage{adjustbox}
\usepackage{makecell}
\usepackage{arydshln}

\usepackage{url}
\usepackage{hyperref}  % 尽量放在最后

\usepackage{booktabs}
\usepackage{adjustbox}
\usepackage{multirow}
\usepackage{arydshln}
\usepackage[table]{xcolor}
\usepackage{colortbl}

\definecolor{bidirectionalbg}{RGB}{222,236,247}   % light blue
\definecolor{autoregressivebg}{RGB}{252,235,213}  % light orange
\definecolor{pairbetterbg}{RGB}{246,199,142}      % medium orange

\newcommand{\pairbetter}[1]{\cellcolor{pairbetterbg}#1}

\definecolor{closedsourcebg}{RGB}{232,232,232}

\newcommand{\best}[1]{\textbf{#1}}
\newcommand{\second}[1]{\underline{#1}}

\title{ProAR: Learning Prospective Reasoning with Autoregressive Video Models}

\author{
\hspace*{-0.3cm}
\begin{tabular}{c}
\textbf{
Linghui Shen$^{\spadesuit}$ \quad
Tinghui Zhu$^{\clubsuit}$ \quad
Sheng Zhang$^{\diamondsuit}$ \quad
Muhao Chen$^{\clubsuit}$
} \\[4pt]
\normalfont
$^{\spadesuit}$ The Hong Kong Polytechnic University
\quad
$^{\clubsuit}$ University of California, Davis
\quad
$^{\diamondsuit}$ Microsoft
\end{tabular}
}

\iclrfinalcopy % Uncomment for camera-ready version, but NOT for submission.
\begin{document}

\maketitle
\lhead{Preprint}

\vspace{-2mm}
\begin{abstract}

Autoregressive (AR) video models excel at causal generation, but their reliance on next-chunk prediction confines them to a short-sighted, reactive paradigm. This limitation is particularly consequential for reasoning-oriented generation, where achieving a target outcome through valid intermediate states matters more than local visual plausibility. 
To address this challenge, we propose Learning \textbf{Pro}spective Reasoning with \textbf{A}uto\textbf{r}egressive Video Models \textbf{(ProAR)}, a novel framework that transforms autoregressive video generation into a goal-oriented reasoning process. ProAR introduces two key components: (1) To anchor generation to the long-range \textit{outcome}, we integrate goal-frame prediction into the autoregressive loop via an asymmetric attention mask, enabling the predicted goal frame to guide the generation of intermediate states without being disrupted by them. (2) To guide short-range \textit{transition}, we introduce future representation self-alignment to encourage current hidden states to anticipate upcoming temporal dynamics. By leveraging teacher-forcing in AR training, we extract clean future representations in a single forward pass and align current representations with them using a lightweight, training-only predictor. Together, these two mechanisms seamlessly combine explicit, sparse target supervision with implicit, dense step-wise guidance, promoting coherent, goal-directed reasoning progress with modest computational cost. Experiments show that ProAR's complementary components consistently improve performance across diverse visual reasoning benchmarks. The framework proves highly training-efficient, surpassing fully trained standard AR baselines using only 25\% of the training steps. This paradigm also demonstrates promising applicability to embodied reasoning tasks. 
Project Page: \url{https://luka-group.github.io/ProAR/}

\end{abstract}
\vspace{-2mm}
\section{Introduction}
\vspace{-2mm}
Video generation models are moving beyond visual synthesis toward reasoning through visual dynamics. 
They have been used to solve game-like problems such as maze navigation and visual puzzles \citep{vbvr,VBVRpro,videorlvr}, simulate physical processes \citep{apple-pi}, and support embodied control \citep{genie,astra,matrix-game2,matrix-game3,dreamzero,lingbot-va}. In these settings, the generated video is itself the reasoning process: the model must achieve a target outcome by generating a sequence of valid visual states. This paradigm requires video models to reason about and plan the evolution of visual states, rather than merely produce visually plausible content, opening new possibilities for physical and embodied intelligence \citep{whitepaper}. We refer to this emerging paradigm as \textit{\textbf{video-native reasoning}}, where a model reasons by directly generating how the visual world evolves.

Autoregressive (AR) video models provide a natural foundation for video-native reasoning.  Unlike bidirectional video  models, AR models generate video sequentially, conditioning each new chunk on the preceding visual history.
This causal generation process supports long-horizon generation and low-latency streaming \citep{Magi-1,causvid,longlive,self-forcing,causal-forcing}, making AR models particularly suitable for real-time and interactive applications \citep{Magi-1,causvid,longlive,self-forcing,causal-forcing}.  

\begin{figure}[t]
    \centering
    \includegraphics[width=0.98\linewidth]{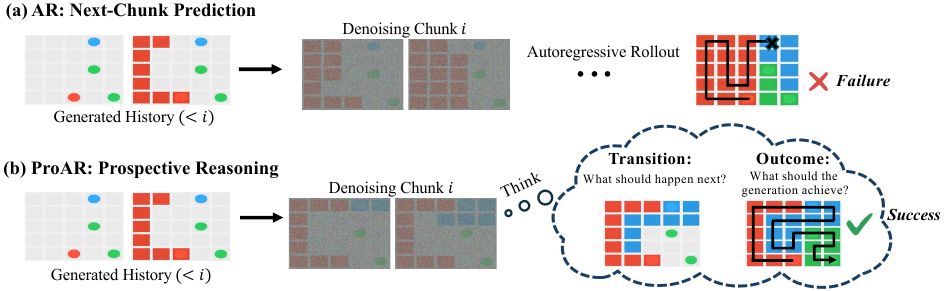}
    \vspace{-3mm}
    \caption{Overview of ProAR. (a) Standard AR relies on local next-chunk prediction, leading to trajectory drift and task failure. (b) ProAR introduces prospective reasoning to anticipate immediate transitions and final outcomes, guiding generation toward task success.}
    \vspace{-4mm}
    \label{fig: overview}
\end{figure}

However, there is a fundamental mismatch between the standard AR training objective and reasoning-oriented generation. Next-chunk prediction optimizes only for locally plausible continuations of the observed history without accounting for their impact on eventual task completion, influencing all subsequent predictions. Consequently, short-sighted decisions compound over time, easily steering a locally plausible trajectory away from the intended goal (Fig.~\ref{fig: overview}(a)). Existing methods primarily focus on improving visual quality, temporal consistency, or prediction horizon \citep{video-mirai,next-forcing,hierarchical,rollingforcing,future-forcing}, rather than explicitly addressing the short-sightedness of AR generation in goal-directed reasoning tasks.

We argue that autoregressive video-native reasoning requires a prospective view of generation.
We define \textit{\textbf{prospective reasoning}} as the capacity to anticipate possible future outcomes and state transitions based on observed history, and to use this foresight to guide ongoing generation.
This perspective exposes two complementary requirements (Fig.~\ref{fig: overview}(b)). At the long-range level, the model should anticipate its target outcome. At the short-range level, it should anticipate upcoming transitions that drive meaningful progress. Long-range outcome awareness provides global direction, while short-range transition awareness encourages coherent and steady progress toward the goal.
Together, they enable future expectations to guide present generation, rather than solely reacting to the past.

Motivated by these requirements, we propose Learning \textbf{Pro}spective Reasoning with \textbf{A}uto\textbf{r}egressive Video Models  (\textbf{ProAR}). ProAR introduces two complementary mechanisms: (1) \textit{Outcome guidance} provides explicit but sparse long-range supervision. At each autoregressive step, ProAR jointly predicts the current chunk and a goal frame representing the anticipated final state. An asymmetric attention mask prevents the current chunk from influencing the goal belief, while allowing the predicted goal to guide current generation. As the video unfolds, the model continuously updates its goal belief based on the growing history. (2) \textit{Transition guidance} provides implicit and dense short-range supervision through future representation self-alignment. Leveraging teacher forcing, ProAR reuses the clean next-chunk representation computed within the same forward pass to supervise the current states to anticipate upcoming temporal dynamics. This design requires neither an external representation encoder nor an additional backbone forward pass, and the lightweight alignment predictor is removed entirely during inference. At inference, ProAR generates autoregressively without access to ground-truth future states, relying on its dynamically updated goal-frame predictions and learned anticipatory representations to guide generation.

% We evaluate ProAR on 13 visual reasoning tasks: 10 perception and spatial reasoning tasks selected from VBVR \citep{vbvr}, and 3 abstract reasoning tasks from VideoRLVR \citep{videorlvr} (Maze, FlowFree, and Sokoban).  Compared with standard AR generation, ProAR improves the average VBVR score by \textbf{XX} and the average VideoRLVR success rate by \textbf{XX}. Outcome and Transition guidance consistently complement each other across these tasks. ProAR also shows strong training efficiency that, on VBVR 10 tasks subset, with only 25\% training steps, it already outperforms the standard AR baseline. Moreover, 

We evaluate ProAR on 13 visual reasoning tasks, including a 10-task subset of VBVR \citep{vbvr} covering perceptual and spatial reasoning, and 3 abstract reasoning tasks from VideoRLVR \citep{videorlvr}. Compared with standard AR generation, ProAR raises the mean score on the 10-task VBVR subset from 0.663 to 0.801, and the average success rate on VideoRLVR from 50.97 to 52.97. Ablation studies further demonstrate that Outcome and Transition guidance provide complementary gains across these tasks. ProAR is also highly training-efficient: on the 10-task VBVR subset, it surpasses the fully trained AR baseline using only 25\% of the training steps. Moreover, ProAR demonstrates promising generalization to embodied reasoning, delivering consistent improvements on the WorldArena simulation benchmark \citep{worldarena}.

Our contributions are threefold: (1) We revisit autoregressive video generation from a reasoning perspective, identifying the lack of prospective guidance as a key limitation of next-chunk prediction. (2) We propose ProAR, which combines goal belief and future representation self-alignment to provide \textit{outcome} and \textit{transition} guidance with minimal overhead. (3) Experiments across diverse video reasoning tasks demonstrate the complementary benefits of outcome and transition, driving consistent improvements in both reasoning performance and training efficiency, with promising generalization to embodied reasoning.

\vspace{-1mm}

\section{Related Work}
\vspace{-2mm}
\paragraph{Autoregressive Video Generation. }
Autoregressive video models generate frames or chunks causally, making them suitable for long, streaming, and interactive generation. Recent methods improve scalability \citep{Magi-1}, efficiency~\citep{longlive, longlive2}, and training stability \citep{causvid,self-forcing,self-forcing++,causal-forcing} during autoregressive rollout.
Several studies have explored richer future information beyond next-chunk prediction through foresight alignment \citep{video-mirai}, hierarchical denoising \citep{hierarchical}, and future-aware training \citep{next-forcing, rollingforcing}. These advances have also enabled interactive world models and world action models for embodied agents \citep{genie,astra,matrix-game3,dreamzero,lingbot-va}. However, extending the prediction horizon alone can improve visual quality, but is insufficient for task-oriented generation which requires both a clear target and effective intermediate transitions. We address these needs with complementary outcome and transition guidance.
\vspace{-2mm}
\paragraph{Native Reasoning with Video Generation Models. }
Recent studies have begun to treat video generation as a paradigm for reasoning rather than only visual synthesis \citep{Thinkwithvideo,VideoZeroshot,VGI-BENCH,MME-CoF-Pro}. Video generators can perform visual reasoning by generating intermediate solution steps, such as navigating a maze or moving pieces to solve a puzzle \citep{vbvr,VBVRpro}. Existing approaches improve reasoning capabilities through visual prompting \citep{Visualprompt}, test-time training \citep{VLMteacher,testtime}, explicit chains of visual thought \citep{Vchain}, and reinforcement learning \citep{videorlvr, VBVRpro}. Other studies further investigate how reasoning capabilities emerge in video generation models \citep{Demystifying, OpenCoF}. However, most of these methods build on bidirectional generators that jointly denoise all frames and allow dense interactions across time. Autoregressive models instead strictly enforce causal generation, committing to each chunk without seeing future frames or revising past errors. The closest work in this setting \citep{hierarchical} introduces hierarchical coarse-to-fine denoising for autoregressive video reasoning. In contrast, ProAR provides lightweight outcome and transition guidance without changing the basic chunk-wise generation process.
\vspace{-2mm}
\paragraph{Representation Learning in Diffusion Models. }
Diffusion models learn rich semantic and structural features in their intermediate layers \citep{dift,Diffusionhyperfeatures}. One line of work aligns generative representations with strong pretrained feature spaces \citep{dinov2,mae,I-jepa} to improve training efficiency and generation quality \citep{Repa,UREPA,Repa-e}. Another line derives supervision from the generative model itself, using stronger internal representations to guide weaker ones \citep{SRA,Layersync,Self-flow}. Recently, representation alignment has also been explored in video post-training \citep{videorepa,video-mirai}. Most relatedly, Video-Mirai \citep{video-mirai} distills future bidirectional representations into autoregressive causal features but requires an additional forward pass. In contrast, our Transition module exploits autoregressive training properties to obtain clean future representations within the model, requiring neither an external representation encoder nor an additional model pass.

\vspace{-2mm}
\section{ProAR}
\label{sec:method}
\vspace{-2mm}
In this section, we present ProAR, a novel framework for learning prospective reasoning in autoregressive video generation. 
We first investigate why local next-chunk supervision is insufficient to ensure goal completion and valid progression (Sec.~\ref{sec: preliminary}). ProAR addresses this gap through two complementary mechanisms. \textit{Outcome guidance} provides explicit, sparse supervision by anchoring generation with an evolving goal belief (Sec.~\ref{sec: outcome}), while \textit{Transition guidance} provides implicit, dense supervision by encouraging every current representation to anticipate the next stage (Sec.~\ref{sec: transition}).

\vspace{-2mm}
\subsection{Preliminaries and Motivation}
\label{sec: preliminary}
\vspace{-2mm}
\paragraph{Autoregressive Video Diffusion.}
In this paper, we consider autoregressive (AR) video diffusion models. Let
$\mathbf{z}^{1:N}=\{\mathbf{z}^{1},\ldots,\mathbf{z}^{N}\}$
denote a video divided into $N$ latent chunks, where
$\mathbf{z}^{<i}$ denotes all chunks preceding $\mathbf{z}^{i}$.
Given a generation condition $\mathbf{c}$, such as an input image and a text prompt, the joint distribution is factorized as: 
\begin{equation}
p_{\theta}(\mathbf{z}^{1:N}\mid\mathbf{c})
=
\prod_{i=1}^{N}
p_{\theta}(\mathbf{z}^{i}\mid\mathbf{z}^{<i},\mathbf{c}).
\end{equation}
Each conditional distribution is modeled with flow matching \citep{flowmatching}. Given a clean target chunk $\mathbf{z}^{i}_{0}$, we construct its noisy state as
$\mathbf{z}^{i}_{t}=(1-t)\mathbf{z}^{i}_{0}+t\boldsymbol{\epsilon}$,
where $\boldsymbol{\epsilon}\sim\mathcal{N}(\mathbf{0},\mathbf{I})$
and $t\in[0,1]$. The velocity network $\mathbf{v}_{\theta}$ is trained with
\begin{equation}
\mathcal{L}_{\mathrm{AR}}
=
\mathbb{E}_{i,t,\boldsymbol{\epsilon}}
\left[
\left\|
\mathbf{v}_{\theta}
\left(
\mathbf{z}^{i}_{t},t
\mid
\mathbf{z}^{<i}_{0},\mathbf{c}
\right)
-
\left(
\boldsymbol{\epsilon}-\mathbf{z}^{i}_{0}
\right)
\right\|_{2}^{2}
\right].
\label{eq: L_ar}
\end{equation}
We adopt teacher-forcing during training. All target chunks are denoised in parallel within a single forward pass, while a causal attention mask ensures each noisy chunk is conditioned only on its clean ground-truth history. During inference, the model generates chunks sequentially and appends each to the history for subsequent prediction. Thus, training provides chunk-level denoising supervision, whereas inference proceeds as an irreversible rollout over model-generated history.
\vspace{-2mm}

\paragraph{From Next-Chunk Prediction to Prospective Reasoning.}
Video-native reasoning requires each generated video to complete a specified task. This imposes two requirements beyond visual plausibility: (1) \textit{goal completion}, where the final state must satisfy the task objective, and (2) \textit{progression validity}, where intermediate transitions must follow task logic or physical constraints.

However, the local objective $\mathcal{L}_{\mathrm{AR}}$ (Eq. \ref{eq: L_ar}) supervises only the current chunk given its preceding history. It does not directly supervise whether the prediction supports the final objective or enables valid subsequent transitions. During inference, every generated chunk is appended to the history and cannot be revised, allowing early errors to persist and compound throughout the autoregressive rollout. As illustrated in Fig.~\ref{fig: overview}(a), a locally plausible rollout may therefore culminate in task failure.

This gap motivates \textit{prospective reasoning} in Fig.~\ref{fig: overview}(b): anticipating state evolution and future outcomes from available observations, then using these expectations to guide current generation. It requires guidance at two temporal scales. Long-range guidance maintains an expectation of the final outcome, while short-range guidance anticipates the next stage of progression. These requirements motivate the outcome and transition mechanisms introduced below.
% As described in Sec.~\ref{preliminary}, a standard AR model learns
% $p_{\theta}(\mathbf{z}^{i}\mid\mathbf{z}^{<i},\mathbf{c})$
% through the chunk-level objective $\mathcal{L}_{\mathrm{AR}}$ during training. Although this objective captures temporal dependencies from the observed history, it evaluates the prediction of the current chunk locally. It provides no direct supervision on whether the predicted state supports eventual task completion or enables valid subsequent transitions. This limitation is amplified during inference, where each generated chunk becomes fixed context for all later predictions. Consequently, an early error can constrain the remaining trajectory and ultimately compromise task completion. As illustrated in Fig.~\ref{background figure}, locally plausible generation may either drift away from the intended goal or violate task rules during intermediate progression.

% This gap motivates a prospective view of AR generation. We define \textit{prospective reasoning} as anticipating future outcomes and state evolution from available observations, and using these expectations to guide current generation. ProAR instantiates this principle at two temporal scales: it maintains an evolving belief about the intended final state (long-range) and encourages current representations to anticipate the next stage (short-range). The following section introduces the corresponding Outcome and Transition mechanisms.

% \muhao{make sure to provide an overview of technical components, better with crossref.}

\begin{figure}[t]
    \centering
    \includegraphics[width=0.98\linewidth]{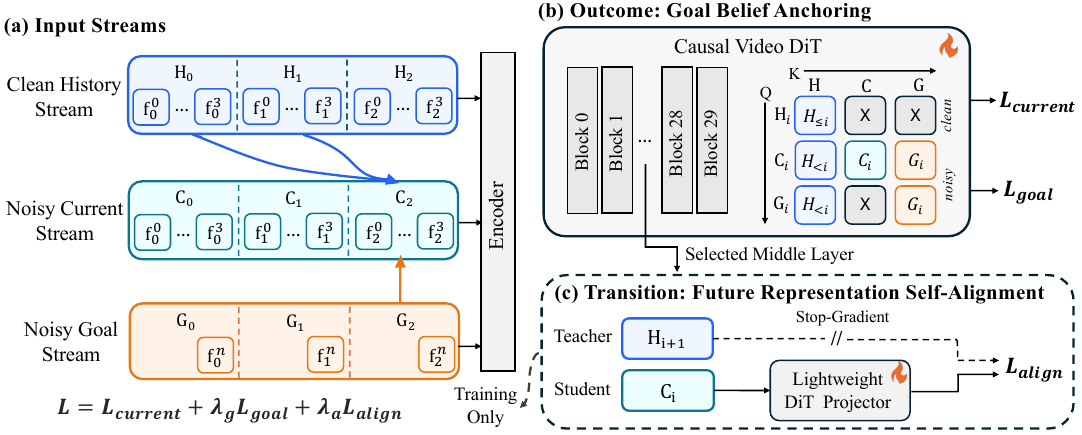}
    \caption{Overview of ProAR. (a) Input: For a given video, the model processes the history, current, and goal streams within a forward pass. (b) Outcome: The causal DiT jointly denoises the current chunk and goal frame, where an attention mask enables the clean history and evolving goal belief to guide current generation. (c) Transition: A training-only projector aligns the middle-layer hidden states of the noisy current chunk (student) with the clean next chunk (teacher).}
    % Both the backbone DiT and the projector are trained.  while preventing the current state from corrupting the goal.
    \vspace{-2mm}
    \label{fig: model}
\end{figure}

\vspace{-2mm}
\subsection{Outcome: Goal Belief for Long-Horizon Planning}
\label{sec: outcome}
Outcome guidance addresses \textit{goal completion} by answering \textit{\textbf{What should the generation achieve?}} It introduces a \textit{goal belief} that explicitly represents the anticipated final state and evolves with the generated history. By jointly predicting the current chunk and goal belief at every AR step, the model can guide local generation toward the expected long-range outcome.

Let $\mathbf{g}^{0}$ denote the clean ground-truth goal frame (final frame). At autoregressive step $i$, the model jointly denoises the noisy current chunk $\mathbf{z}_{i}^{t}$ and the noisy goal input $\mathbf{g}_{i}^{t}$, conditioned on the clean history $\mathbf{z}_{<i}^{0}$ and generation condition $\mathbf{c}$. Both inputs are constructed at the same flow-matching timestep $t$:
\begin{equation}
\mathbf{z}_{i}^{t}
=
(1-t)\mathbf{z}_{i}^{0}
+
t\boldsymbol{\epsilon}_{z,i},
\qquad
\mathbf{g}_{i}^{t}
=
(1-t)\mathbf{g}^{0}
+
t\boldsymbol{\epsilon}_{g,i},
\end{equation}
where
$\boldsymbol{\epsilon}_{z,i},\boldsymbol{\epsilon}_{g,i}
\sim\mathcal{N}(\mathbf{0},\mathbf{I})$
are sampled independently. A shared backbone predicts the velocity fields for both streams in a single forward pass. The current stream is supervised by the AR objective $\mathcal{L}_{\mathrm{current}} = \mathcal{L}_{\mathrm{AR}}$ in Eq.~\ref{eq: L_ar}, while the goal stream is supervised by
\begin{equation}
\mathcal{L}_{\mathrm{goal}}
=
\mathbb{E}_{i,t,\boldsymbol{\epsilon}_{g,i}}
\left[
\left\|
\mathbf{v}_{\theta}^{g}
\left(
\mathbf{g}_{i}^{t},t
\mid
\mathbf{z}_{<i}^{0},\mathbf{c}
\right)
-
\left(
\boldsymbol{\epsilon}_{g,i}-\mathbf{g}^{0}
\right)
\right\|_{2}^{2}
\right],
\label{eq:goal-loss}
\end{equation}
where $\mathbf{v}_{\theta}^{g}$ denotes the velocity predicted for the goal stream.

Joint prediction alone cannot guarantee the desired information flow. We therefore introduce an asymmetric attention mask (Fig.~\ref{fig: model}(b)). Specifically, goal tokens attend only to the clean history and themselves, shielded from the noisy current chunk. Conversely, current tokens additionally attend to these goal tokens. This target-first dependency prevents the uncertain current state from corrupting the goal belief, while allowing the anticipated goal to guide ongoing generation, modulated by the loss weight $\lambda_{\mathrm{goal}}$ in Eq. \ref{eq: Loss}.

During training, the goal branch at every AR step is supervised by the same ground-truth final frame $\mathbf{g}^{0}$. During inference, the model jointly denoises the current chunk and goal input, but only the generated current chunk is appended to the history. The goal belief is re-estimated at the next step from the expanded history, allowing it to be progressively refined as the video unfolds.

\vspace{-1mm}
\subsection{Transition: Representation Alignment for Progressive Reasoning}
\label{sec: transition}
\vspace{-1mm}
Outcome guidance specifies where the generation should end, but leaves how the current state should evolve unconstrained. Transition guidance addresses \textit{progression validity} by answering \textit{\textbf{What should happen next?}} Our key intuition is that the current representation should anticipate the next stage of the task. We realize this intuition through \textit{future representation self-alignment}.
% \vspace{-2mm}

\vspace{-1mm}
\paragraph{Teacher and Student.}
Recent studies show that intermediate hidden states of diffusion models encode rich semantic and temporal abstractions, and that stronger hidden states can guide weaker ones to improve generation \citep{SRA,Layersync}. Building on this insight, we construct a self-supervised Teacher--Student mechanism within the backbone feature space.

Our core idea is to use the clean representation of the next chunk to guide the noisy representation of the current chunk, as illustrated in Fig.~\ref{fig: model}(c). The clean next-chunk hidden states contain uncontaminated information about the next stage and therefore serve as a strong, prospective \textit{teacher}. In contrast, the noisy current hidden states contain incomplete and noise-corrupted information, serving as the weaker \textit{student}. Thus, the teacher provides not only a cleaner representation but also a short-range temporal direction for the current state.

Formally, at a selected backbone layer $\ell$, let $\mathbf{H}_{i+1}^{\ell}$ and $\mathbf{C}_{i}^{t,\ell}$ denote the hidden states of the clean next chunk $\mathbf{z}_{i+1}^{0}$ and the noisy current chunk $\mathbf{z}_{i}^{t}$, respectively (simplified as $\mathbf{H}_{i+1}$ and $\mathbf{C}_{i}$ in Fig.~\ref{fig: model}(c)). For each valid current chunk ($i<N$), a lightweight predictor $\mathcal{P}_{\phi}$ maps the student toward the fixed teacher representation:
\begin{equation}
\overline{\mathbf{H}}_{i+1}^{\ell}
=
\operatorname{sg}
\left(
\mathbf{H}_{i+1}^{\ell}
\right),
\qquad
\widehat{\mathbf{H}}_{i+1}^{t,\ell}
=
\mathcal{P}_{\phi}
\left(
\mathbf{C}_{i}^{t,\ell}
\right),
\end{equation}
where $\operatorname{sg}$ denotes the stop-gradient operator. The predictor (default 3 transformer blocks) learns to transform the student features to anticipate future dynamics, while preserving the backbone's token grid and dimension. This ensures direct spatial alignment without dimensional compression.

\vspace{-2mm}
\paragraph{Single-Pass Self-Supervision.}
A key advantage of our design is that both Teacher and Student hidden states are obtained from the same backbone forward pass. During teacher-forcing training, all noisy current chunks are denoised in parallel, each attending only to its clean ground-truth history (Fig.~\ref{fig: model}(a)). 
Consequently, the clean next chunk $\mathbf{z}_{i+1}^{0}$ is inherently encoded in the history stream to condition later denoising targets, while the noisy current chunk $\mathbf{z}_i^{t}$ is processed simultaneously in the current stream.
We thus obtain their hidden states $\mathbf{H}_{i+1}^{\ell}$ and $\mathbf{C}_i^{t,\ell}$ entirely for free, eliminating the overhead of an external representation encoder or an additional backbone pass.

Motivated by evidence that reasoning capabilities primarily localize in intermediate DiT blocks \citep{Demystifying}, we extract these representations from a shared middle layer. We empirically validate this structural choice through layer-wise ablations in Appendix~\ref{app: ablation2}.
% where middle-layer alignment yields the strongest reasoning performance and produces activations that focus more clearly on task-relevant entities and transition paths.
\vspace{-2mm}

\paragraph{Representation Alignment.}
We align the student's prediction $\widehat{\mathbf{H}}_{i+1}^{t,\ell}$ with the teacher target $\overline{\mathbf{H}}_{i+1}^{\ell}$ using a token-wise cosine distance loss.  Given $M$ tokens per chunk, the alignment objective is:
\begin{equation}
\mathcal{L}_{\mathrm{align}}
=
\mathbb{E}_{i<N,t}
\left[
\frac{1}{M}
\sum_{m=1}^{M}
\left(
1-
\cos
\left(
\widehat{\mathbf{H}}_{i+1,m}^{t,\ell},
\overline{\mathbf{H}}_{i+1,m}^{\ell}
\right)
\right)
\right].
\end{equation}
The stop-gradient operation keeps the teacher fixed, while gradients propagate through the predictor $\mathcal{P}_{\phi}$ and the student branch of the backbone. This directional optimization encourages the current hidden states to anticipate upcoming temporal dynamics.
% The loss is applied only for $i<N$, since the final chunk has no valid successor.
\vspace{-2mm}
\paragraph{Overall Objective and Inference.}
The complete training objective combines current generation, goal prediction (Sec. \ref{sec: outcome}), and future representation alignment (Sec. \ref{sec: transition}):
\begin{equation}
\mathcal{L}
=
\mathcal{L}_{\mathrm{current}}
+
\lambda_{\mathrm{goal}}\mathcal{L}_{\mathrm{goal}}
+
\lambda_{\mathrm{align}}\mathcal{L}_{\mathrm{align}}.
\label{eq: Loss}
\end{equation}
At inference, the alignment predictor $\mathcal{P}_{\phi}$ is removed entirely, so Transition guidance adds no inference cost. Outcome guidance only requires jointly denoising 1 auxiliary goal frame with each current chunk in the same forward pass. Detailed computational costs are provided in Appendix~\ref{app: implementation}.

\begin{figure}[t]
    \centering
    \includegraphics[width=0.98\linewidth]{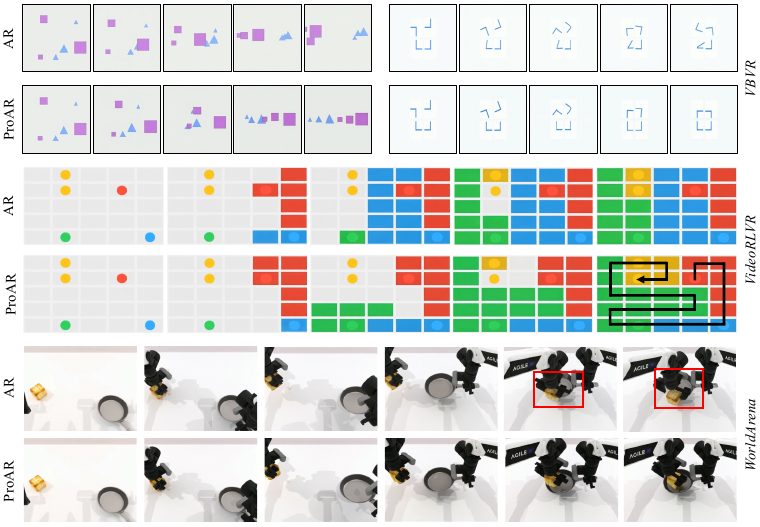}
    \caption{Qualitative Results. Across tasks from VBVR, VideoRLVR, and WorldArena, ProAR consistently improves reasoning performance by promoting goal completion and valid progression. }
    \vspace{-4mm}
    \label{fig: qualitative}
\end{figure}

\vspace{-2mm}
\section{Experiments}

\subsection{Experimental Setup}
\label{sec: exp}
\noindent\textbf{Benchmarks.}
We evaluate ProAR on three complementary video-native reasoning benchmarks.

\textit{(1) VBVR.}
VBVR \citep{vbvr} contains 100 tasks, each with 10K training examples. We select 10 tasks that primarily assess perceptual and spatial reasoning: Concentric Ring (G194), Outline Match (G25), Stable Sort (G3), Star Match (G5), Domino Chain (O23), Ball Eater (O31), Grid Shift (O36), Pipe Puzzle (O44), Slide Puzzle (O47), and Balancing Vessels (O75). Since the official benchmark provides only five test examples per task, we use its official data generator to expand the test set to 50 examples per task, yielding 500 test examples in total.

\textit{(2) VideoRLVR.}
VideoRLVR \citep{videorlvr} evaluates abstract reasoning through 3 games: Maze, FlowFree, and Sokoban. Each task contains 10K training and 1K test samples. Together, VBVR and VideoRLVR cover 13 tasks spanning perceptual, spatial, and abstract reasoning.

\textit{(3) WorldArena.}
As an extension to embodied reasoning, we evaluate ProAR on WorldArena \citep{worldarena}, following its official protocol on 50 bimanual manipulation tasks from RoboTwin 2.0 \citep{robotwin2.0}. Each task provides 50 training and 20 separately generated test samples. This setting tests whether ProAR can generate actionable sequences as an embodied video simulator.

\begin{table*}[t]
\centering
\caption{
Performance comparison across 10 selected tasks in VBVR.
The best result for each metric is shown in \best{bold},
and the second-best result is \second{underlined}.
}
\label{tab:vbvr_results}

\renewcommand{\arraystretch}{1.18}
\setlength{\tabcolsep}{4pt}

\begin{adjustbox}{max width=\textwidth}
\begin{tabular}{lccccccccccc}
    \toprule

    \textbf{Model}
    & \makecell{\textbf{Concentric}\\\textbf{Ring}}
    & \makecell{\textbf{Outline}\\\textbf{Match}}
    & \makecell{\textbf{Stable}\\\textbf{Sort}}
    & \makecell{\textbf{Star}\\\textbf{Match}}
    & \makecell{\textbf{Domino}\\\textbf{Chain}}
    & \makecell{\textbf{Ball}\\\textbf{Eater}}
    & \makecell{\textbf{Grid}\\\textbf{Shift}}
    & \makecell{\textbf{Pipe}\\\textbf{Puzzle}}
    & \makecell{\textbf{Slide}\\\textbf{Puzzle}}
    & \makecell{\textbf{Balancing}\\\textbf{Vessels}}
    & \textbf{Mean} \\
    \midrule

    % Bidirectional methods

\multicolumn{12}{l}{%
    \cellcolor{closedsourcebg}%
    \footnotesize\textbf{\textit{Closed-Source Models}}%
} \\
    Seedance 2.0
    & 0.653
    & 0.658
    & 0.693
    & 0.598
    & 0.547
    & 0.696
    & 0.482
    & \second{0.530}
    & 0.799
    & 0.616
    & 0.627 \\
    \multicolumn{12}{l}{%
        \cellcolor{bidirectionalbg}%
        \footnotesize\textbf{\textit{Bidirectional Models}}%
    } \\
    Wan2.2 5B
    & 0.241
    & 0.480
    & 0.400
    & 0.312
    & 0.419
    & 0.349
    & 0.000
    & 0.256
    & 0.034
    & 0.362
    & 0.285 \\

    Wan2.2 5B SFT
    & 0.798
    & 0.702
    & \best{0.905}
    & 0.827
    & 0.521
    & 0.779
    & 0.585
    & 0.514
    & 0.860
    & 0.872
    & 0.736 \\

    % Autoregressive methods
    \multicolumn{12}{l}{%
        \cellcolor{autoregressivebg}%
        \footnotesize\textbf{\textit{Autoregressive Models}}%
    } \\

    Standard AR
    & 0.794
    & 0.753
    & 0.565
    & 0.689
    & \best{0.658}
    & 0.754
    & 0.720
    & 0.424
    & 0.388
    & \second{0.880}
    & 0.663 \\

    \hdashline

    w/ Outcome
    & \second{0.940}
    & \second{0.755}
    & 0.890
    & \second{0.858}
    & 0.639
    & \best{0.784}
    & \second{0.726}
    & 0.486
    & \second{0.908}
    & 0.869
    & \second{0.786} \\

    w/ Transition
    & 0.860
    & 0.758
    & 0.699
    & 0.659
    & \second{0.646}
    & 0.765
    & 0.651
    & 0.348
    & 0.558
    & \best{0.887}
    & 0.683 \\

    \textbf{ProAR (ours)}
    & \best{0.949}
    & \best{0.777}
    & \second{0.895}
    & \best{0.871}
    & 0.628
    & \second{0.781}
    & \best{0.770}
    & \best{0.546}
    & \best{0.931}
    & 0.863
    & \best{0.801} \\

    \bottomrule
\end{tabular}
\end{adjustbox}
\vspace{-4mm}
\end{table*}

\begin{table*}[t]
    \centering
    \caption{
        Ablation study on the Transition training start point.
        (* denotes default.)
    }
    \label{tab:training_start_point_ablation}

    \renewcommand{\arraystretch}{1.16}
    \setlength{\tabcolsep}{4pt}

    \begin{adjustbox}{max width=\textwidth}
        \begin{tabular}{lccccccccccc}
            \toprule

            \textbf{Start Point}
            & \makecell{\textbf{Concentric}\\\textbf{Ring}}
            & \makecell{\textbf{Outline}\\\textbf{Match}}
            & \makecell{\textbf{Stable}\\\textbf{Sort}}
            & \makecell{\textbf{Star}\\\textbf{Match}}
            & \makecell{\textbf{Domino}\\\textbf{Chain}}
            & \makecell{\textbf{Ball}\\\textbf{Eater}}
            & \makecell{\textbf{Grid}\\\textbf{Shift}}
            & \makecell{\textbf{Pipe}\\\textbf{Puzzle}}
            & \makecell{\textbf{Slide}\\\textbf{Puzzle}}
            & \makecell{\textbf{Balancing}\\\textbf{Vessels}}
            & \textbf{Mean} \\

            \midrule
            Step 0
            & 0.909
            & \second{0.771}
            & \second{0.909}
            & 0.850
            & 0.566
            & \best{0.784}
            & \best{0.782}
            & \best{0.591}
            & \best{0.978}
            & 0.857
            & \second{0.799} \\
    
            Step 5000
            & \second{0.913}
            & 0.755
            & \best{0.923}
            & \second{0.870}
            & \second{0.610}
            & 0.766
            & 0.741
            & 0.423
            & \second{0.953}
            & \best{0.864}
            & 0.782 \\
    
            Step 7500$^{\ast}$
            & \best{0.949}
            & \best{0.777}
            & 0.895
            & \best{0.871}
            & \best{0.628}
            & \second{0.781}
            & \second{0.770}
            & \second{0.546}
            & 0.931
            & \second{0.863}
            & \best{0.801} \\
        
            \bottomrule
        \end{tabular}
    \end{adjustbox}
    \vspace{-5mm}
\end{table*}

\begin{table*}[t]
    \centering
    \caption{
        Performance comparison across 3 tasks in VideoRLVR.
        $\dagger$ are results reported from VideoRLVR \citep{videorlvr}.
    }
    \label{tab:videorlvr_results}

    \renewcommand{\arraystretch}{1.16}
    \setlength{\tabcolsep}{4pt}

    \begin{adjustbox}{width=0.95\textwidth,center}
        \begin{tabular}{lcccccccccccc}
            \toprule

            \multirow{2}{*}{\textbf{Model}}
            & \multicolumn{4}{c}{\textbf{Maze}}
            & \multicolumn{4}{c}{\textbf{FlowFree}}
            & \multicolumn{4}{c}{\textbf{Sokoban}} \\

            \cmidrule(lr){2-5}
            \cmidrule(lr){6-9}
            \cmidrule(lr){10-13}

            & \textbf{Prec.}
            & \textbf{Rec.}
            & \textbf{F1}
            & \textbf{SR}
            & \textbf{Prec.}
            & \textbf{Rec.}
            & \textbf{F1}
            & \textbf{SR}
            & \textbf{Prec.}
            & \textbf{Rec.}
            & \textbf{F1}
            & \textbf{SR} \\

            \midrule

            % Bidirectional models
            \multicolumn{13}{l}{
                \cellcolor{bidirectionalbg}
                \footnotesize\textbf{\textit{Bidirectional Models}}
            } \\

            Wan2.2 SFT$^\dagger$
            & 80.40
            & 85.10
            & 82.70
            & 69.00
            & 43.10
            & 42.50
            & 42.80
            & 2.50
            & 32.80
            & 11.60
            & 17.10
            & 2.70 \\

            VideoRLVR$^\dagger$
            & 82.10
            & 86.90
            & 84.40
            & 72.20
            & 44.30
            & 43.80
            & 44.00
            & 7.90
            & 34.00
            & 12.50
            & 29.40
            & 6.10 \\

            % Autoregressive models
            \multicolumn{13}{l}{
                \cellcolor{autoregressivebg}
                \footnotesize\textbf{\textit{Autoregressive Models}}
            } \\

        Standard AR
        & 89.28
        & 91.86
        & 90.16
        & 79.50
        & 55.22
        & 54.91
        & 55.05
        & \second{29.40}
        & 40.18
        & \second{34.74}
        & 34.62
        & 44.00 \\

        \hdashline

        w/ Outcome
        & \second{93.01}
        & \best{96.54}
        & \best{94.65}
        & \second{82.30}
        & \second{57.83}
        & 55.16
        & \second{56.10}
        & 23.20
        & 43.63
        & 32.73
        & 35.59
        & 44.30 \\

        w/ Transition
        & 89.29
        & 91.99
        & 90.15
        & 80.90
        & 55.80
        & \second{55.43}
        & 55.59
        & 26.70
        & \second{45.41}
        & \best{34.93}
        & \second{36.65}
        & \best{49.60} \\

        \textbf{ProAR (ours)}
        & \best{93.33}
        & \second{95.88}
        & \second{94.48}
        & \best{82.70}
        & \best{60.57}
        & \best{58.86}
        & \best{59.52}
        & \best{30.20}
        & \best{47.62}
        & 32.56
        & \best{36.87}
        & \second{46.00} \\

            \bottomrule
        \end{tabular}
    \end{adjustbox}
\vspace{-4mm}
\end{table*}

\noindent\textbf{Baselines.} 
For each benchmark, we evaluate both bidirectional and autoregressive video generation.

\textit{(1) Bidirectional Models:} On the VBVR, we evaluate both the pretrained Wan2.2-TI2V-5B model and its SFT variant trained following the official protocol \citep{vbvr}. On VideoRLVR, we use the reported Wan2.2-TI2V-5B results after supervised fine-tuning and reinforcement learning.

\textit{(2) Autoregressive Models:} We adapt the open-source Wan2.2-TI2V-5B \citep{wan2.2} backbone to autoregressive generation using causal attention and teacher-forcing training. This Standard AR baseline is trained and evaluated under the same settings as ProAR.

\noindent\textbf{Implementation Details.}
We optimize all models using AdamW with a learning rate of $5\times10^{-6}$ and a global batch size of 16. The chunk size is 4 frames for VBVR and VideoRLVR, and 8 frames for WorldArena. We conduct every training on 2-4 NVIDIA B200 GPUs.

For \textit{Outcome Guidance}, the final ground-truth frame serves as the goal target. For \textit{Transition Guidance}, we extract both student (noisy current chunk) and teacher (clean next chunk) hidden states from backbone block 15 in a single forward pass. A lightweight 3-block DiT predictor maps the Student to the Teacher representation and is fully removed at inference.

We set $\lambda_{\mathrm{goal}}=0.25$ and $\lambda_{\mathrm{align}}=0.01$ by default. Since the raw cosine alignment loss is much larger than the generation losses, the weighted alignment term contributes approximately 3\%--10\% of the total loss. Transition guidance is activated in the later training stages: at step 7,500/10,000 for VBVR and step 3,000/5,000 for VideoRLVR. See more details in Appendix~\ref{app: implementation}.
% Additional implementation details, ablations and metrics illustrations are provided in Appendix~\ref{app: implementation}.

\subsection{Main Results}

\paragraph{Quantitative Results on VBVR.}
Table~\ref{tab:vbvr_results} summarizes performance across the selected 10 perceptual and spatial reasoning tasks. Consistent with the short-horizon limitation discussed in Sec.~\ref{sec: preliminary}, the teacher-forced Standard AR baseline achieves a mean score of 0.663, below the 0.736 obtained by bidirectional Wan2.2 SFT. With prospective guidance, ProAR raises the mean score to 0.801, a 20.8\% relative improvement over Standard AR. It also exceeds the closed-source Seedance 2.0 and Wan2.2 SFT baselines by 0.174 and 0.065, respectively. At the task level, ProAR outperforms Standard AR on 8 of the 10 tasks, with particularly large absolute gains on Stable Sort (+0.330), Pipe Puzzle (+0.122), and Slide Puzzle (+0.543). The gains across tasks with different objectives indicate that the improvement is not driven by a single reasoning pattern. Overall, these results show that prospective guidance effectively reduces the mismatch between local chunk prediction and reasoning task completion.

\vspace{-2mm}
\paragraph{Quantitative Results on VideoRLVR.}
Table~\ref{tab:videorlvr_results} reports results on the 3 abstract reasoning games. Surprisingly, Standard AR already outperforms both bidirectional Wan2.2 SFT and its reinforcement-learning variant, suggesting that causal generation is well suited to structured tasks involving sequential state transitions. Building on this strong baseline, ProAR improves the average success rate from 50.97 to 52.97 and the average precision from 61.56 to 67.17. Precision increases on all 3 games, including a substantial gain from 40.18 to 47.62 on Sokoban. ProAR also improves the success rate over Standard AR on every game, reaching 82.70 on Maze, 30.20 on FlowFree, and 46.00 on Sokoban. These consistent gains demonstrate the effectiveness of prospective guidance across different abstract reasoning environments.

\vspace{-2mm}
\paragraph{Qualitative Results.}
As illustrated in Fig.~\ref{fig: qualitative}, ProAR overcomes the short-sighted failures of standard AR by consistently ensuring valid progression and successful goal completion. On VBVR, AR produces disorganized movements and incomplete shapes, whereas ProAR correctly aligns objects to form the intended patterns. On VideoRLVR, standard AR generates fragmented or invalid paths, while ProAR produces continuous, rule-consistent trajectories that reach the target.

\subsection{Training Efficiency}
\vspace{-1mm}
\begin{wrapfigure}{r}{0.4\textwidth}
    \centering
    \vspace{-3mm}
    \includegraphics[width=\linewidth]{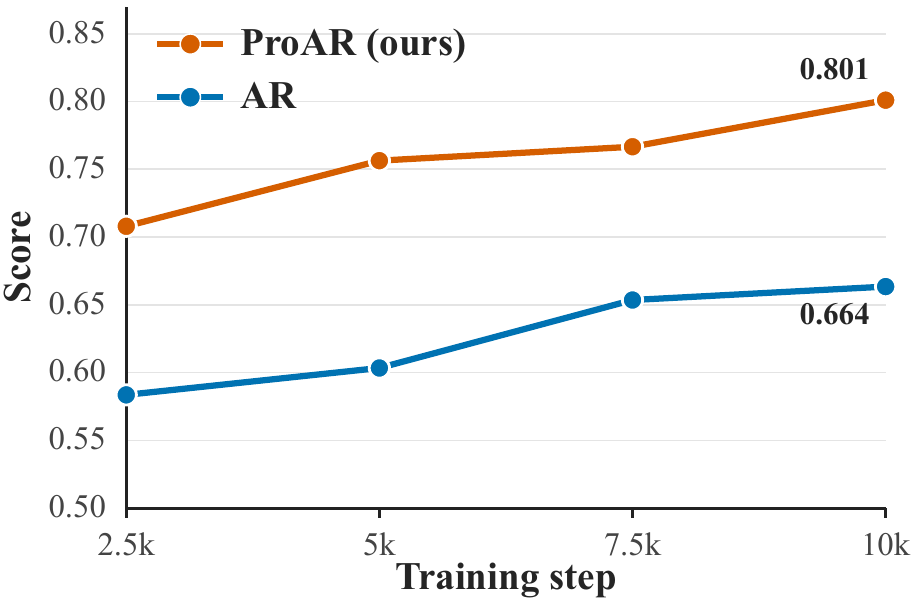}
    \vspace{-7mm}
    \caption{Performance curve on VBVR 10 tasks subsets.}
    \label{fig: efficiency}
\end{wrapfigure}
We next examine whether prospective guidance improves training efficiency
in addition to final performance.
Fig. \ref{fig: efficiency} compares ProAR with the standard AR baseline at matched training checkpoints on VBVR 10-task subset. 
ProAR converges substantially faster. After only 2,500 steps, it reaches
a mean score of 0.708, already exceeding the 0.663 obtained by the
standard AR baseline after 10,000 steps. After Transition guidance is
activated at step 7,500, the score further increases from 0.7666 to
0.8010 by step 10,000. Overall, these results show that prospective
guidance improves both convergence speed and final reasoning performance
without requiring additional optimization steps.

\subsection{Ablations and Analysis}
\vspace{-2mm}

\label{sec: ablation1}

\paragraph{Complementarity of Outcome and Transition.}
To isolate the contribution of each component, we evaluate ProAR with only outcome guidance (\textit{w/ Outcome}) or transition guidance (\textit{w/ Transition}) in Tables~\ref{tab:vbvr_results} and~\ref{tab:videorlvr_results}. On VBVR, outcome and transition individually improve the mean score of Standard AR from 0.663 to 0.786 and 0.683, respectively, while their combination further increases it to 0.801. A similar trend appears on VideoRLVR: Outcome and Transition raise the average precision from 61.56 to 64.82 and 63.50, whereas the complete ProAR reaches 67.17. Overall, ProAR achieves the best aggregate performance on both benchmarks, leading among the AR variants on 7 of the 10 VBVR tasks and 8 of the 12 VideoRLVR metrics. These results show that long-range outcome awareness and short-range transition anticipation provide distinct and complementary guidance.

\vspace{-3mm}

\paragraph{When to Activate Transition Alignment.}
We investigate when to activate the transition module during training in Table~\ref{tab:training_start_point_ablation}. Applying transition alignment throughout training from scratch (Step 0) yields comparable or slightly inferior performance to activating it at a later stage (Step 7500). We hypothesize that, early in training, the backbone has not yet developed stable representations for the target reasoning domain, making its internal teacher targets less informative. A task-specific fine-tuning warm-up allows more reliable, task-relevant structure to emerge before self-alignment begins. We therefore adopt the late-start strategy as the default configuration in all main experiments.

Ablations on the attention mask, transition layer, and predictor depth are provided in Appendix~\ref{app: ablation2}.

\subsection{Extension on Embodied Reasoning}
\label{sec: extension1}
We further evaluate whether ProAR extends beyond visual reasoning to embodied manipulation. Table~\ref{tab:worldarena_selected_results} reports selected WorldArena results, with the complete evaluation and metrics illustration provided in Appendix~\ref{app:worldarena-results}. ProAR achieves an EWMScore of 60.95, the second-highest result among the methods shown, remaining highly competitive with computationally heavier bidirectional baselines and specialized embodied world models.

Under the controlled comparison with Standard AR, ProAR improves 8 of the 9 component metrics and raises EWMScore from 60.43 to 60.95. The gains are particularly evident in task-relevant dimensions, including Depth Accuracy (88.42 to 91.11), Instruction Following (80.88 to 81.96), Interaction Quality (73.12 to 73.86), and Trajectory Accuracy (41.74 to 42.34). ProAR also improves motion and background consistency while maintaining comparable visual quality. These results suggest that prospective guidance transfers effectively to embodied settings, helping AR video models generate physically coherent and task-directed manipulation trajectories. As shown in Fig. \ref{fig: qualitative},  on WorldArena, standard AR rollouts gradually drift and lead to failed grasps, whereas ProAR preserves stable physical interactions and completes the manipulation successfully.

\begin{table*}[t]
\centering
\caption{
Extension study on WorldArena.
Within the last two rows, the higher result for each metric
is highlighted in orange.
}
\label{tab:worldarena_selected_results}

\renewcommand{\arraystretch}{1.05}
\setlength{\tabcolsep}{3.5pt}

\begin{adjustbox}{width=0.93\textwidth,center}
\begin{tabular}{l*{10}{c}}
\toprule

\textbf{Method}
& \makecell{\textbf{Image}\\\textbf{Qual.}}
& \makecell{\textbf{JEPA}\\\textbf{Sim.}}
& \makecell{\textbf{Flow}\\\textbf{Score}}
& \makecell{\textbf{Back.}\\\textbf{Cons.}}
& \makecell{\textbf{Photo.}\\\textbf{Cons.}}
& \makecell{\textbf{Interact.}\\\textbf{Qual.}}
& \makecell{\textbf{Traj.}\\\textbf{Acc.}}
& \makecell{\textbf{Depth}\\\textbf{Acc.}}
& \makecell{\textbf{Instr.}\\\textbf{Follow.}}
& \textbf{EWMScore}  \\

\midrule

% Bidirectional models
\multicolumn{11}{l}{%
    \cellcolor{bidirectionalbg}%
    \footnotesize\textbf{\textit{Bidirectional Models}}%
} \\

CogVideoX
& 36.23
& 94.84
& \second{21.77}
& 88.38
& 33.79
& 66.96
& 34.79
& 91.09
& 75.58
& 58.79 \\

Cosmos-Predict 2.5 (text)
& \best{65.49}
& 46.69
& 17.55
& 73.98
& 35.39
& 65.14
& 11.60
& 70.71
& 64.70
& 53.06 \\

Cosmos-Predict 2.5 (action)
& 49.48
& 90.65
& 9.95
& 72.72
& \second{45.21}
& 60.46
& 27.49
& 90.35
& 60.54
& 54.29 \\

Vidar
& 41.75
& 61.10
& 17.03
& 78.24
& 20.57
& 60.84
& 21.26
& 79.77
& 61.54
& 51.92 \\

FlowWAM
& 48.87
& \best{97.14}
& 20.47
& \second{89.97}
& 29.78
& \second{73.82}
& \best{64.26}
& \best{98.97}
& \best{82.76}
& \best{63.71} \\

% Autoregressive models
\multicolumn{11}{l}{%
    \cellcolor{autoregressivebg}%
    \footnotesize\textbf{\textit{Autoregressive Models}}%
} \\

IRASim
& 41.85
& 93.72
& 11.66
& 78.69
& \best{45.70}
& 62.76
& 35.92
& \second{93.50}
& 65.40
& 56.15 \\

Genie Envisioner
& 28.18
& 33.81
& 19.91
& 87.54
& 16.04
& 22.74
& 2.63
& 86.83
& 20.36
& 41.37 \\

Ctrl-World
& 42.44
& 92.77
& \best{33.57}
& \best{90.30}
& 12.88
& 62.62
& \second{48.20}
& 93.25
& 67.68
& 59.98 \\

\hdashline

Standard AR
& 50.01
& 95.30
& 19.73
& 87.10
& \pairbetter{23.17}
& 73.12
& 41.74
& 88.42
& 80.88
& 60.43 \\

\textbf{ProAR (ours)}
& \pairbetter{\second{50.23}}
& \pairbetter{\second{95.34}}
& \pairbetter{20.71}
& \pairbetter{88.11}
& 23.12
& \pairbetter{\best{73.86}}
& \pairbetter{42.34}
& \pairbetter{91.11}
& \pairbetter{\second{81.96}}
& \pairbetter{\second{60.95}} \\

\bottomrule
\end{tabular}
\end{adjustbox}
\vspace{-3mm}
\end{table*}

% 1. vbvr 10 tasks, + ablation (bidirectional, ar, module 1, module 2, module 1+2)

% 2. videorlvr 10 tasks, + ablation (bidirectional, ar, module 1, module 2, module 1+2)

% 3. ablation on vbvr 10 tasks:  (appendix)
% module 1: asymmetric attention mask 
% module 2: dit1,2,3; transformer block 10, 15, 20; training from 5000 to 10000, 7500 to 10000

% 4. improve training efficiency: vbvr 10 tasks, AR vs D$^2$AR, 2500/5000/7500/10000

% 5. extension data domain: physics reasoning, gui, embodied ai

% 6. computational cost report: training \& inference

\section{Conclusion}

In this paper, we present ProAR, a novel framework that equips autoregressive video generation with prospective reasoning capability. To address the short-sightedness of next-chunk prediction, ProAR combines outcome guidance for goal completion with transition guidance for valid intermediate progression. Experiments across perceptual, spatial, abstract, and embodied reasoning benchmarks demonstrate consistent improvements with modest computational overhead.
These results highlight prospective modeling as a promising direction for advancing video-native reasoning, opening new possibilities for physical and
embodied intelligence.

\subsection*{AI Use Statement}
Generative AI tools supported manuscript review, experimental
planning, code development, and language editing.
The authors conducted the experiments, verified the code,
results, text, and references, and take full responsibility
for the final manuscript.

\subsection*{Ethics Statement}
This work develops prospective reasoning methods for video
generation, evaluated on established benchmarks and simulated
manipulation tasks. The research aims to improve goal-directed
prediction and understanding of video models. We identify no
specific ethical concerns arising from the proposed methodology
or the reported experimental setting.

\subsection*{Reproducibility Statement}
Section~\ref{sec:method} describes the autoregressive formulation,
goal conditioning, and transition alignment objectives.
Section~\ref{sec: exp} and Appendix~\ref{app: implementation}
specify datasets, training and inference settings, and
computational costs. Section~\ref{sec: ablation1} and
Appendix~\ref{app: ablation2} provide ablation studies.
Appendices~\ref{app: task_spec} and~\ref{app:worldarena-results}
detail the reasoning-task evaluation protocols and
WorldArena metrics.

\newpage
\bibliography{iclr2027_conference}
\bibliographystyle{iclr2027_conference}

\newpage
\appendix
\section{Appendix}
In this supplement, we provide implementation and computational details (Sec.\ref{app: implementation}), additional ablations of the Outcome and Transition designs (Sec.\ref{app: ablation2}), visual analyses of AR failure modes and future representation alignment (Sec.\ref{sec: analysis}), task specifications and evaluation protocols (Sec.\ref{app: task_spec}), complete WorldArena results (Sec.\ref{app:worldarena-results}), a discussion of limitation (Sec.\ref{app: discussion}), and further qualitative examples (Sec.\ref{app: visuals}).

\subsection{Implementation Details}
\label{app: implementation}

\paragraph{Video Configuration.}
In addition to the chunk settings detailed in the main paper, VBVR videos are generated at $512\times512$ resolution with task-specific sequence lengths. VideoRLVR videos consist of 81 frames at $832\times480$ resolution. During inference, we employ 20 denoising steps per chunk. Further ablations on the configuration settings are detailed in Sec.~\ref{app: ablation2}.

\paragraph{WorldArena Settings.}
WorldArena \citep{worldarena} evaluates video world models as embodied simulators across 50 RoboTwin 2.0 tasks \citep{robotwin2.0}. Since the original trajectory lengths vary substantially, we follow the FlowWAM \citep{flowwam} protocol to standardize the training data: each video is uniformly sampled to 121 frames and resized to $640\times480$ resolution. For evaluation, WorldArena introduces 1,000 newly generated test trajectories (20 per task) beyond the official RoboTwin dataset. Following WorldArena's default configuration, we generate these test videos in the identical 121-frame and $640\times480$ format. This ensures structural consistency between training and inference while preserving the complete physical manipulation process. Given these longer videos, we use a chunk size of 8 and set $\lambda_{\mathrm{goal}}=0.25$ and $\lambda_{\mathrm{align}}=0.05$. We train for 10000 steps and start the transition module at step 7000.

\paragraph{Computational Cost.}
ProAR introduces only modest computational overhead over Standard AR during both training and inference.

\textit{Training:} Outcome guidance appends one goal frame to each 4-frame chunk, increasing the token count by 8\%--11\% for videos of 3--8 chunks. Transition guidance is highly efficient: it reuses internal representations from the same forward pass without requiring an external teacher. Its lightweight 3-block predictor (236M parameters, just 4.7\% of the 5B backbone) adds a mere 2\%--3\% to the training computation. Overall, ProAR increases the total training cost by only 12\%--14\%. 

\textit{Inference:} The Transition predictor is discarded entirely, yielding zero overhead. Meanwhile, the Outcome module jointly denoises the goal frame alongside the current chunk without any additional forward passes or sampling steps. Measured on an NVIDIA B200 GPU, ProAR increases the overall inference time by approximately 12\% for a chunk size of 4 compared to standard AR generation. Since only a single goal frame is added per step, this relative overhead diminishes further with larger chunk sizes. For example, generating 8-frame chunks in WorldArena incurs a marginal time increase of only $\sim$8\%.

\subsection{Additional Ablation Studies}
\label{app: ablation2}
In this section, we provide additional ablations on the attention mask, alignment layer, and predictor depth, complementing the studies in Sec.~\ref{sec: ablation1}.

\paragraph{Outcome: Asymmetric Attention Mask.}
We first validate the asymmetric attention design introduced in Sec.~\ref{sec: outcome}. As shown in Table~\ref{tab: ablation_mask}, removing the mask allows unrestricted interaction between the noisy current chunk and the noisy goal tokens. Our asymmetric design achieves a slightly higher mean score, improving from 0.799 to 0.801, and performs better on 6 of the 10 tasks. Although the overall difference is modest, the result supports the proposed information flow: the goal belief is protected from the uncertain current state while remaining available to guide current generation. Moreover, the asymmetric attention pattern is sparser, making it more computationally efficient. We therefore use the asymmetric mask as the default Outcome configuration.
\begin{table*}[!h]
    \centering
    \caption{
        Ablation on asymmetric attention mask.
    }
    \label{tab: ablation_mask}

    \renewcommand{\arraystretch}{1.16}
    \setlength{\tabcolsep}{4pt}

    \begin{adjustbox}{max width=\textwidth}
        \begin{tabular}{lccccccccccc}
            \toprule

            \textbf{Model}
            & \makecell{\textbf{Concentric}\\\textbf{Ring}}
            & \makecell{\textbf{Outline}\\\textbf{Match}}
            & \makecell{\textbf{Stable}\\\textbf{Sort}}
            & \makecell{\textbf{Star}\\\textbf{Match}}
            & \makecell{\textbf{Domino}\\\textbf{Chain}}
            & \makecell{\textbf{Ball}\\\textbf{Eater}}
            & \makecell{\textbf{Grid}\\\textbf{Shift}}
            & \makecell{\textbf{Pipe}\\\textbf{Puzzle}}
            & \makecell{\textbf{Slide}\\\textbf{Puzzle}}
            & \makecell{\textbf{Balancing}\\\textbf{Vessels}}
            & \textbf{Mean} \\

            \midrule

            w/o Asym. Mask
            & 0.911
            & 0.750
            & \textbf{0.913}
            & 0.865
            & 0.620
            & \textbf{0.797}
            & 0.742
            & \textbf{0.588}
            & \textbf{0.950}
            & 0.861
            & 0.799 \\
    
            \textbf{ProAR}
            & \textbf{0.949}
            & \textbf{0.777}
            & 0.895
            & \textbf{0.871}
            & \textbf{0.628}
            & 0.781
            & \textbf{0.770}
            & 0.546
            & 0.931
            & \textbf{0.863}
            & \textbf{0.801} \\

            \bottomrule
        \end{tabular}
    \end{adjustbox}
\end{table*}

\paragraph{Transition: Alignment Layer.}
We next study where future representation self-alignment should be applied within the backbone. Table~\ref{tab: ablation_layer} compares Layers 10, 15, and 20, which span the middle portion of the 30-block DiT. \underline{All three settings} outperform both the standard AR baseline (0.663; Standard AR in Table~\ref{tab:vbvr_results}) and Outcome guidance alone (0.786; w/ Outcome in Table~\ref{tab:vbvr_results}), showing that Transition guidance is effective across a range of intermediate layers. Layer 15 achieves the highest mean score of 0.801, compared with 0.788 and 0.791 for Layers 10 and 20, respectively. We therefore align the Student and Teacher hidden states at Layer 15 by default.

\begin{table*}[!h]
    \centering
    \caption{
        Ablations on the choice of alignment layer. (* denotes default.)
    }
    \label{tab: ablation_layer}

    \renewcommand{\arraystretch}{1.16}
    \setlength{\tabcolsep}{4pt}

    \begin{adjustbox}{max width=\textwidth}
        \begin{tabular}{lccccccccccc}
            \toprule

            \textbf{Alignment Layer}
            & \makecell{\textbf{Concentric}\\\textbf{Ring}}
            & \makecell{\textbf{Outline}\\\textbf{Match}}
            & \makecell{\textbf{Stable}\\\textbf{Sort}}
            & \makecell{\textbf{Star}\\\textbf{Match}}
            & \makecell{\textbf{Domino}\\\textbf{Chain}}
            & \makecell{\textbf{Ball}\\\textbf{Eater}}
            & \makecell{\textbf{Grid}\\\textbf{Shift}}
            & \makecell{\textbf{Pipe}\\\textbf{Puzzle}}
            & \makecell{\textbf{Slide}\\\textbf{Puzzle}}
            & \makecell{\textbf{Balancing}\\\textbf{Vessels}}
            & \textbf{Mean} \\

            \midrule

            Layer 10
            & \second{0.948}
            & 0.763
            & 0.868
            & 0.866
            & \best{0.647}
            & \best{0.785}
            & 0.646
            & \best{0.563}
            & \second{0.928}
            & \best{0.867}
            & 0.788 \\
    
            Layer 15$^{\ast}$
            & \best{0.949}
            & \best{0.777}
            & \second{0.895}
            & \best{0.871}
            & 0.628
            & \second{0.781}
            & \second{0.770}
            & \second{0.546}
            & \best{0.931}
            & \second{0.863}
            & \best{0.801} \\
    
            Layer 20
            & 0.944
            & \second{0.767}
            & \best{0.907}
            & \second{0.867}
            & \second{0.643}
            & 0.764
            & \best{0.773}
            & 0.483
            & 0.902
            & 0.862
            & \second{0.791} \\

            \bottomrule
        \end{tabular}
    \end{adjustbox}
\end{table*}

\paragraph{Transition: Predictor Depth.}
Finally, we ablate the depth of the lightweight DiT projector in Table~\ref{tab: ablation_ditnum}. Increasing its depth from 2 to 3 blocks improves the mean score from 0.792 to 0.801. This suggests that the additional block provides greater capacity to map the noisy current hidden states to the clean next-chunk representation. Since the predictor is used only during training, the 3-block design improves alignment without increasing inference cost. We adopt it as our default configuration.

\begin{table*}[!h]
    \centering
    \caption{
        Ablation study on the number of projector DiTs.
        ($^{\ast}$ denotes default.)
    }
    \label{tab: ablation_ditnum}

    \renewcommand{\arraystretch}{1.16}
    \setlength{\tabcolsep}{4pt}

    \begin{adjustbox}{max width=\textwidth}
        \begin{tabular}{lccccccccccc}
            \toprule

            \textbf{DiT \#}
            & \makecell{\textbf{Concentric}\\\textbf{Ring}}
            & \makecell{\textbf{Outline}\\\textbf{Match}}
            & \makecell{\textbf{Stable}\\\textbf{Sort}}
            & \makecell{\textbf{Star}\\\textbf{Match}}
            & \makecell{\textbf{Domino}\\\textbf{Chain}}
            & \makecell{\textbf{Ball}\\\textbf{Eater}}
            & \makecell{\textbf{Grid}\\\textbf{Shift}}
            & \makecell{\textbf{Pipe}\\\textbf{Puzzle}}
            & \makecell{\textbf{Slide}\\\textbf{Puzzle}}
            & \makecell{\textbf{Balancing}\\\textbf{Vessels}}
            & \textbf{Mean} \\

            \midrule

            % 1 DiT
            % & -- & -- & -- & -- & --
            % & -- & -- & -- & -- & -- & -- \\

            2 DiTs
            & 0.932
            & 0.770
            & 0.860
            & 0.865
            & 0.609
            & \best{0.785}
            & \best{0.791}
            & 0.510
            & \best{0.933}
            & \best{0.867}
            & 0.792 \\
    
            3 DiTs$^{\ast}$
            & \best{0.949}
            & \best{0.777}
            & \best{0.895}
            & \best{0.871}
            & \best{0.628}
            & 0.781
            & 0.770
            & \best{0.546}
            & 0.931
            & 0.863
            & \best{0.801} \\

            \bottomrule
        \end{tabular}
    \end{adjustbox}
\end{table*}

\subsection{Analysis}
\label{sec: analysis}
We provide visual analyses that complement the motivation in Sec.~\ref{sec: preliminary} and examine the Transition mechanism introduced in Sec.~\ref{sec: transition}. Specifically, we illustrate the characteristic failure modes of Standard AR and visualize how future representation alignment shifts current hidden states toward the next video stage.

\begin{figure}[!h]
    \centering
    \includegraphics[width=0.95\linewidth]{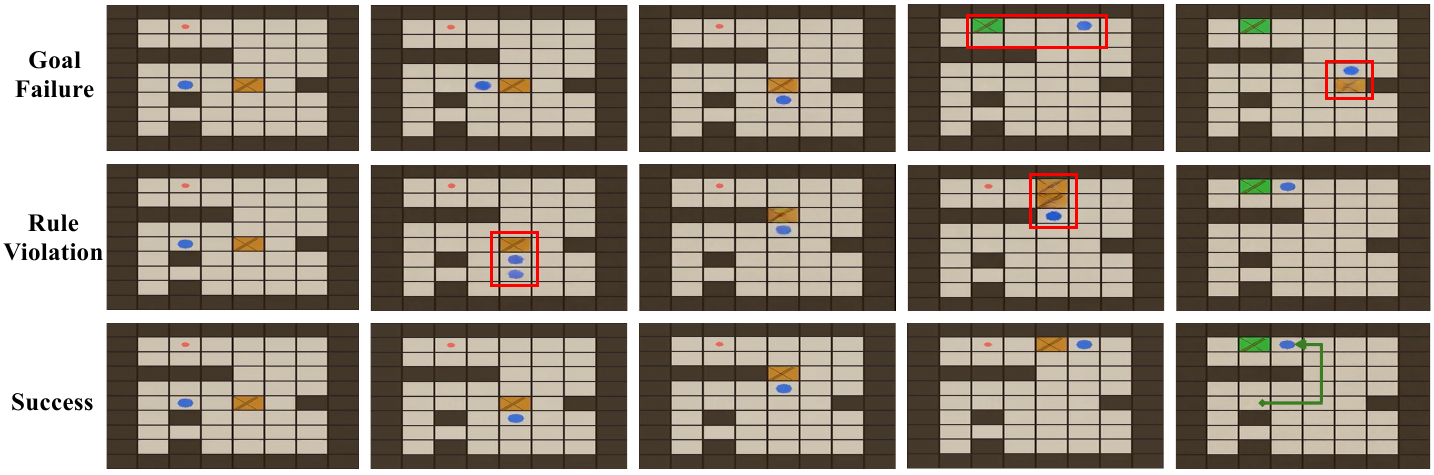}
    \caption{Failure Modes of Standard AR. }
    \label{fig: failure_modes}
\end{figure}

\paragraph{Failure Modes of Standard AR.}
As motivated in Sec.~\ref{sec: preliminary}, video-native reasoning requires both goal completion and valid progression. Fig.~\ref{fig: failure_modes} illustrates how Standard AR can fail either requirement: a rollout may remain locally plausible but end without completing the task (\textit{Goal Failure}), or produce invalid intermediate interactions that violate task rules (\textit{Rule Violation}). In contrast, a successful trajectory preserves valid transitions and reaches the intended goal. Red boxes highlight erroneous states, while green arrows indicate the successful path.

\paragraph{Visualization of Future Representation Alignment.}
As described in Sec.~\ref{sec: transition}, transition guidance injects prospective information into current hidden states. To examine this effect, we visualize the resulting representations. During teacher-forcing training, all video chunks are processed in parallel within a single forward pass. For each transition from chunk $i$ to chunk $i+1$, we extract the noisy current hidden states from Layer 15 at a moderate noise level ($\sigma=0.5$). We then pass them through the trained lightweight DiT predictor to obtain the student prediction. The clean hidden states of chunk $i+1$, extracted from the clean video stream at the same layer, serve as the stop-gradient teacher target. We project the current, predicted, and target representations into RGB using a shared PCA basis and normalization range, allowing their spatial structures to be compared.

\begin{figure}[!h]
    \centering
    \includegraphics[width=0.95\linewidth]{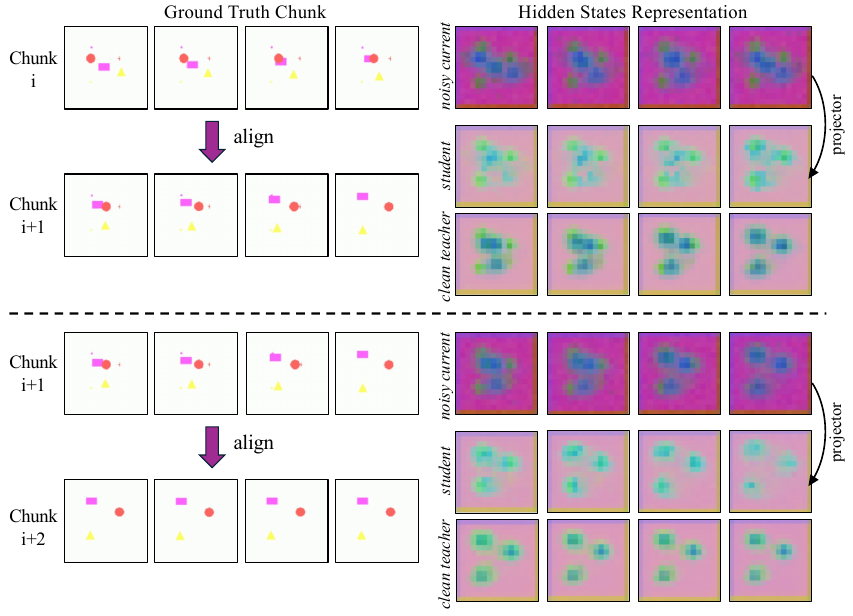}
    \caption{PCA Visualization of Future Representation Alignment.}
    \label{fig: analysis1}
\end{figure}

As shown in Fig.~\ref{fig: analysis1}, the noisy current representations primarily reflect the spatial state of the current chunk. After prediction, the Student representations shift toward the corresponding clean Teacher representations, recovering the spatial layout and object positions of the next chunk. This pattern is observed for both transitions, from $i$ to $i+1$ and from $i+1$ to $i+2$. The visualization suggests that the predictor does not merely reconstruct the current state. Instead, it learns to map current hidden states toward the upcoming video state, encouraging the backbone to encode prospective features that support progressive reasoning.

\subsection{Reasoning Tasks and Evaluation Protocols}
\label{app: task_spec}
This section details the reasoning tasks and evaluation protocols used for VBVR and VideoRLVR.

\paragraph{Task Specifications.}
Tables~\ref{tab:vbvr_task_specs} and~\ref{tab:videorlvr_task_specs}
summarize the reasoning tasks used in our experiments.

% Insert Tables~\ref{tab:vbvr_task_specs} and
% \ref{tab:videorlvr_task_specs} here.

\paragraph{Evaluation Metrics.}
We follow the official evaluation protocols of VBVR and VideoRLVR.

For VBVR \citep{vbvr}, task-specific rule-based evaluators assess the generated video against its reference solution and task objective. Depending on the task, they measure final-state correctness, object and spatial consistency, transition validity, and rule compliance. Each video receives a continuous score in $[0,1]$, where a higher value indicates greater task completion. This score should not be interpreted as a binary success rate.

For VideoRLVR \citep{videorlvr}, we report Precision, Recall, F1, and Success Rate (SR), macro-averaged over the test samples. F1 is the harmonic mean of the per-sample Precision and Recall. The task-specific metrics are defined as follows. For \textit{Maze}, Precision and Recall measure the overlap between the generated and reference path regions, while SR requires a connected path from start to goal without corrupting the maze walls. For \textit{FlowFree}, Precision and Recall measure the color correctness and coverage of filled non-endpoint cells. SR additionally requires valid non-overlapping paths, preserved endpoints, correct color connectivity, and sufficient grid coverage. For \textit{Sokoban}, Precision and Recall compare the generated and reference action sequences at corresponding steps. SR requires all boxes to reach their targets through sufficiently valid state transitions.

Precision, Recall, and F1 measure local agreement with the reference solution, whereas SR evaluates strict task completion. Consequently, a high F1 score does not necessarily imply a high success rate.

\begin{table*}[t]
    \centering
    \caption{Specifications of the 10 selected VBVR tasks \citep{vbvr}.}
    \label{tab:vbvr_task_specs}
    \renewcommand{\arraystretch}{1.12}
    \setlength{\tabcolsep}{6pt}
    \small
    \begin{tabular}{p{0.25\textwidth} p{0.68\textwidth}}
        \toprule
        \textbf{Task} & \textbf{Description} \\
        \midrule
        Concentric Ring (G194)
        & Move two circles to the same center while preserving their sizes and colors. \\

        Outline Match (G25)
        & Move each colored shape into its matching outline. \\

        Stable Sort (G3)
        & Group shapes by type and sort each group from smallest to largest. \\

        Star Match (G5)
        & Move each colored object to the star marker of the same color. \\

        Domino Chain (O23)
        & Trigger the first domino and propagate the falling sequence through the complete chain. \\

        Ball Eater (O31)
        & Guide the black ball to consume smaller balls, growing after each step. \\

        Grid Shift (O36)
        & Shift all blocks by the specified distance and direction. \\

        Pipe Puzzle (O44)
        & Rotate the pipe tiles until they form a continuous connected path. \\

        Slide Puzzle (O47)
        & Slide tiles into the empty space to sort them. \\

        Balancing Vessels (O75)
        & Simulate liquid flow until all communicating vessels reach the same level. \\
        \bottomrule
    \end{tabular}
\end{table*}

\begin{table*}[t]
    \centering
    \caption{Specifications of the 3 VideoRLVR tasks \citep{videorlvr}.}
    \label{tab:videorlvr_task_specs}
    \renewcommand{\arraystretch}{1.12}
    \setlength{\tabcolsep}{6pt}
    \small
    \begin{tabular}{p{0.25\textwidth} p{0.68\textwidth}}
        \toprule
        \textbf{Task} & \textbf{Description} \\
        \midrule
        Maze
        & Move the agent from the green start to the red goal through open cells without crossing any walls. \\

        FlowFree
        & Connect each pair of same-colored endpoints with a continuous path, without overlapping other paths, until the entire grid is filled. \\

        Sokoban
        & Move the agent through free cells and push the box onto the green target without crossing walls or pulling the box. \\
        \bottomrule
    \end{tabular}
\end{table*}

\subsection{Additional WorldArena Evaluation}
\label{app:worldarena-results}

This section supplements Sec.~\ref{sec: extension1} with detailed metric definitions, descriptions of the evaluated baselines, and complete quantitative results.

\begin{table*}[t]
\centering
\caption{
Performance comparison on WorldArena.
Within the last two rows, the higher result for each metric
is highlighted in orange.
}
\label{tab:worldarena_results}

\renewcommand{\arraystretch}{1.10}
\setlength{\tabcolsep}{3pt}
% Suggested group colors
% \definecolor{bidirectionalbg}{RGB}{214,234,226}
% \definecolor{undisclosedbg}{RGB}{235,235,235}
% \definecolor{blockautoregressivebg}{RGB}{231,224,242}
% \definecolor{autoregressivebg}{RGB}{247,226,205}

\begin{adjustbox}{max width=\textwidth}
\begin{tabular}{l*{17}{c}}
\toprule

\multirow{2}{*}{\textbf{Method}}
& \multicolumn{3}{c}{\textbf{Visual Quality}}
& \multicolumn{3}{c}{\textbf{Motion Quality}}
& \multicolumn{3}{c}{\textbf{Content Consistency}}
& \multicolumn{2}{c}{\textbf{Physics}}
& \multicolumn{2}{c}{\textbf{3D Accuracy}}
& \multicolumn{3}{c}{\textbf{Controllability}}
& \multirow{2}{*}{\textbf{EWMScore}} \\

\cmidrule(lr){2-4}
\cmidrule(lr){5-7}
\cmidrule(lr){8-10}
\cmidrule(lr){11-12}
\cmidrule(lr){13-14}
\cmidrule(lr){15-17}

& \textbf{IQ}
& \textbf{AQ}
& \textbf{JEPA}
& \textbf{DD}
& \textbf{FS}
& \textbf{MS}
& \textbf{SC}
& \textbf{BC}
& \textbf{PC}
& \textbf{IntQ}
& \textbf{TA}
& \textbf{DA}
& \textbf{Per}
& \textbf{InsF}
& \textbf{SA}
& \textbf{AcF}
& \\

\midrule

% Bidirectional models
\multicolumn{18}{l}{%
    \cellcolor{bidirectionalbg}%
    \footnotesize\textbf{\textit{Bidirectional Models}}%
} \\

CogVideoX
& 36.23
& 36.30
& 94.84
& 31.47
& \second{21.77}
& 73.30
& 80.97
& 88.38
& 33.79
& 66.96
& 34.79
& 91.09
& 85.46
& 75.58
& 89.70
& 0.00
& 58.79 \\

Cosmos-Predict 2.5 (text)
& \best{65.49}
& \best{44.33}
& 46.69
& 21.62
& 17.55
& 70.93
& 66.19
& 73.98
& 35.39
& 65.14
& 11.60
& 70.71
& \best{98.22}
& 64.70
& 86.34
& \best{10.01}
& 53.06 \\

Cosmos-Predict 2.5 (action)
& 49.48
& 34.14
& 90.65
& 18.96
& 9.95
& 65.66
& 67.24
& 72.72
& \second{45.21}
& 60.46
& 27.49
& 90.35
& 84.68
& 60.54
& \second{89.86}
& 1.31
& 54.29 \\

Vidar
& 41.75
& 39.87
& 61.10
& 21.05
& 17.03
& \best{79.11}
& 70.37
& 78.24
& 20.57
& 60.84
& 21.26
& 79.77
& 81.88
& 61.54
& 88.46
& \second{7.86}
& 51.92 \\

FlowWAM
& 48.87
& \second{40.83}
& \best{97.14}
& 30.41
& 20.47
& 71.65
& \second{82.46}
& \second{89.97}
& 29.78
& \second{73.82}
& \best{64.26}
& \best{98.97}
& 94.56
& \best{82.76}
& \best{89.93}
& 3.50
& \best{63.71} \\

% Autoregressive models
\multicolumn{18}{l}{%
    \cellcolor{autoregressivebg}%
    \footnotesize\textbf{\textit{Autoregressive Models}}%
} \\

IRASim
& 41.85
& 32.86
& 93.72
& 21.60
& 11.66
& 64.31
& 72.82
& 78.69
& \best{45.70}
& 62.76
& 35.92
& \second{93.50}
& 84.16
& 65.40
& 89.43
& 3.94
& 56.15 \\

Genie Envisioner
& 28.18
& 26.39
& 33.81
& \second{32.41}
& 19.91
& 69.17
& 74.53
& 87.54
& 16.04
& 22.74
& 2.63
& 86.83
& 54.32
& 20.36
& 85.98
& 1.07
& 41.37 \\

Ctrl-World
& 42.44
& 37.05
& 92.77
& \best{41.82}
& \best{33.57}
& \second{77.34}
& \best{83.56}
& \best{90.30}
& 12.88
& 62.62
& 48.20
& 93.25
& 83.66
& 67.68
& 88.68
& 3.90
& 59.98 \\

% GigaWorld-1
% & \second{51.18}
% & \second{41.17}
% & \second{96.77}
% & 30.52
% & 18.64
% & 68.33
% & 80.97
% & 86.43
% & 28.11
% & \best{75.10}
% & \second{54.27}
% & \second{98.44}
% & \second{95.60}
% & \second{82.14}
% & 89.42
% & 0.28
% & \second{62.34} \\

\hdashline

Standard AR
& 50.01
& \pairbetter{39.92}
& 95.30
& 31.31
& 19.73
& 69.74
& 79.90
& 87.10
& \pairbetter{23.17}
& 73.12
& 41.74
& 88.42
& 94.80
& 80.88
& 89.65
& \pairbetter{2.07}
& 60.43 \\

\textbf{ProAR (ours)}
& \pairbetter{\second{50.23}}
& 39.65
& \pairbetter{\second{95.34}}
& \pairbetter{31.38}
& \pairbetter{20.71}
& \pairbetter{69.95}
& \pairbetter{80.86}
& \pairbetter{88.11}
& 23.12
& \pairbetter{\best{73.86}}
& \pairbetter{42.34}
& \pairbetter{91.11}
& \pairbetter{\second{95.04}}
& \pairbetter{\second{81.96}}
& \pairbetter{89.71}
& 1.88
& \pairbetter{\second{60.95}} \\

% % Closed-source models
% \multicolumn{18}{l}{%
%     \cellcolor{closedsourcebg}%
%     \footnotesize\textbf{\textit{Closed-Source Models}}%
% } \\

% Veo 3.1
% & \underline{65.57}
% & \textbf{48.79}
% & 57.97
% & 14.66
% & 8.26
% & 67.85
% & 75.82
% & \textbf{91.67}
% & 37.52
% & \underline{81.50}
% & 11.36
% & 74.27
% & \textbf{99.64}
% & \textbf{97.14}
% & 83.79
% & 8.52
% & 57.77 \\

\bottomrule
\end{tabular}
\end{adjustbox}
\end{table*}

\paragraph{Metrics.}
We follow the official WorldArena evaluation protocol, which includes 16 metrics across 6 dimensions. All metrics are higher-is-better.

\begin{itemize}
    \item \textit{Visual Quality.}
    Image Quality (IQ) measures frame-level perceptual quality using MUSIQ, while Aesthetic Quality (AQ) evaluates overall visual appeal. JEPA Similarity (JEPA) compares high-level spatiotemporal features between generated and reference videos using V-JEPA.

    \item \textit{Motion Quality.}
    Dynamic Degree (DD) measures salient motion in active regions, and Flow Score (FS) captures the overall magnitude of optical flow. Motion Smoothness (MS) evaluates whether motion remains continuous and free from abrupt changes or temporal flickering.

    \item \textit{Content Consistency.}
    Subject Consistency (SC) measures whether robots and objects preserve their appearance across frames. Background Consistency (BC) evaluates scene stability, while Photometric Consistency (PC) measures local appearance consistency after optical-flow alignment.

    \item \textit{Physics.}
    Interaction Quality (IntQ) uses a vision-language model to assess whether robot--object interactions are physically plausible, including contact, grasping, and object motion. Trajectory Accuracy (TA) compares the generated and reference end-effector trajectories after detecting and aligning the robot gripper.

    \item \textit{3D Accuracy.}
    Depth Accuracy (DA) compares the depth structure of generated and reference videos. Perspectivity (Per) evaluates whether scale, occlusion, spatial relationships, and perspective remain geometrically plausible.

    \item \textit{Controllability.}
    Instruction Following (InsF) evaluates whether the generated video correctly executes the given instruction. Semantic Alignment (SA) measures agreement with the reference video at the task-semantic level. Action Following (AcF) evaluates whether different action conditions produce distinguishable behaviors, rather than directly measuring task success.
    \item \textit{EWMScore.}
    WorldArena summarizes overall performance using EWMScore, computed as the arithmetic mean of the 16 metrics after normalizing each to $[0,100]$:
    \begin{equation}
    \mathrm{EWMScore}
    =
    \frac{1}{16}
    \sum_{k=1}^{16} s_k .
    \end{equation}
    It provides a unified measure of visual fidelity, motion, content consistency, physical plausibility, 3D structure, and controllability.
    
\end{itemize}

\paragraph{Baselines.}
We compare ProAR with general video generators, embodied world models, and a controlled autoregressive baseline. We use the results reported from FlowWAM \citep{flowwam}.

\noindent\textit{General Video Generators.}
\begin{itemize}
    \item \textit{CogVideoX.}
    CogVideoX \citep{Cogvideox} is a latent video diffusion model built on an expert transformer and a temporally compressed 3D VAE.

    \item \textit{Cosmos-Predict 2.5.}
    Cosmos-Predict 2.5 \citep{Cosmos-Predict2.5} is a flow-based world foundation model that unifies text-, image-, and video-conditioned generation. We evaluate both its text-conditioned and action-conditioned variants.
\end{itemize}

\noindent\textit{Embodied World Models.}
\begin{itemize}
    \item \textit{IRASim.}
    IRASim \citep{Irasim} is an action-conditioned latent diffusion transformer that predicts future robot observations from historical frames and action trajectories.

    \item \textit{Vidar.}
    Vidar \citep{vidar} combines large-scale multi-view video diffusion pretraining with masked inverse dynamics to infer robot actions from predicted visual trajectories.

    \item \textit{Genie Envisioner.}
    Genie Envisioner \citep{GenieEnvisioner} uses an instruction-conditioned video diffusion model for autoregressive robotic simulation and policy learning.

    \item \textit{Ctrl-World.}
    Ctrl-World \citep{Ctrl-world} combines joint multi-view prediction, frame-level action conditioning, and pose-conditioned memory retrieval to support controllable and temporally consistent rollouts.

    \item \textit{FlowWAM.}
    FlowWAM \citep{flowwam} represents actions as dense optical-flow videos and jointly models flow and RGB observations using a dual-stream diffusion framework.
\end{itemize}

\noindent\textit{Controlled Autoregressive Baseline.}
\begin{itemize}
    \item \textit{Standard AR. }
    We adapt the Wan2.2-TI2V-5B \citep{wan2.2} backbone to autoregressive generation using causal attention and teacher-forcing training. This serves as our Standard AR baseline and follows the same training and evaluation protocol as ProAR.
\end{itemize}

\paragraph{Results.}
As shown in Table~\ref{tab:worldarena_results}, ProAR achieves an EWMScore of 60.95, ranking first among the autoregressive models and second across all evaluated systems. More importantly, under the controlled comparison with standard AR, ProAR improves 13 of the 16 base metrics and raises EWMScore from 60.43 to 60.95. The gains span several task-relevant dimensions, including Depth Accuracy (88.42 to 91.11), Instruction Following (80.88 to 81.96), Background Consistency (87.10 to 88.11), Interaction Quality (73.12 to 73.86), and Trajectory Accuracy (41.74 to 42.34). These results indicate that prospective guidance improves not only visual consistency but also spatial accuracy, physical interaction, and task alignment, supporting the extension of ProAR from visual reasoning to embodied video simulation.

\subsection{Limitations and Discussions}
\label{app: discussion}
While ProAR establishes a powerful paradigm for video-native reasoning, our current implementation of Outcome guidance relies on a simple heuristic: utilizing the final video frame to represent the target outcome. This design choice is highly effective for most goal-oriented reasoning tasks, where the final frame naturally encapsulates the goal completion state.

However, this rigid assumption introduces a limitation. In scenarios where the final frame is semantically uninformative, such as tasks with post-completion idle states or purely process-driven dynamics, predicting this frame may provide weak supervision. Consequently, the model may struggle to extract meaningful long-range guidance.

A promising direction for future work is to transition from a fixed final-frame target to adaptive milestone guidance. Rather than strictly anchoring to the last frame, future frameworks could dynamically detect the most salient keyframes or critical bottleneck states within the visual sequence, utilizing them as flexible goals or intermediate subgoals. This would further generalize prospective reasoning to more open-ended and complex physical environments.

\subsection{Qualitative Results}
\label{app: visuals}
\begin{figure}[!h]
    \centering
    \includegraphics[width=0.85\linewidth]{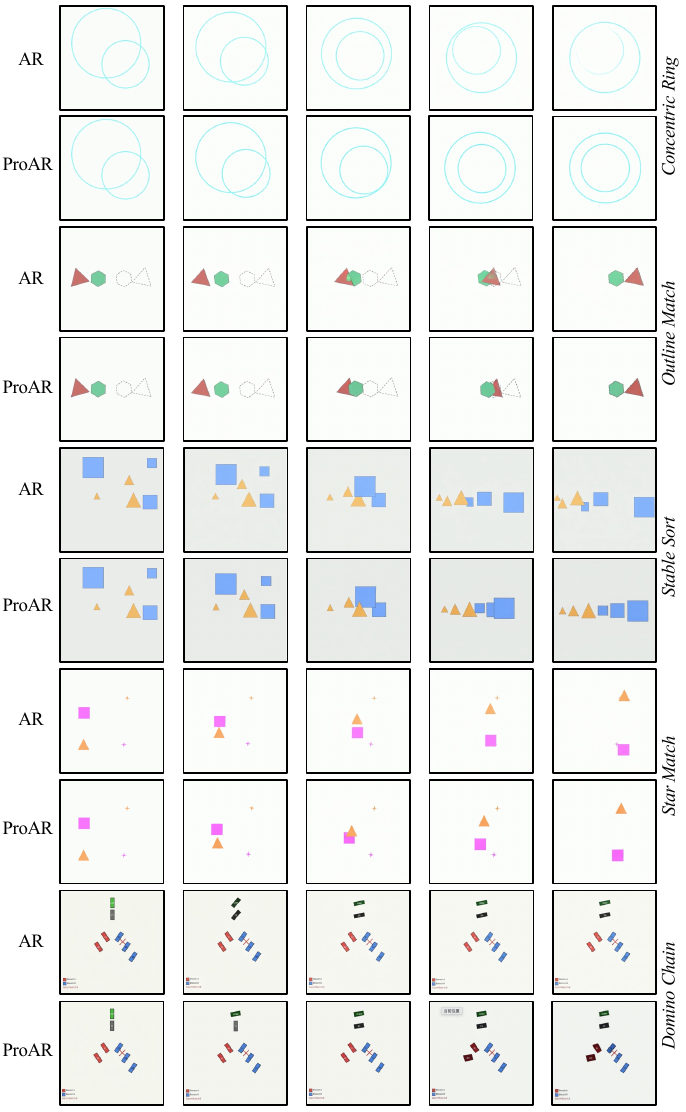}
    \caption{Qualitative Results on VBVR. }
    \label{fig: vbvr1}
\end{figure}

\begin{figure}[!h]
    \centering
    \includegraphics[width=0.85\linewidth]{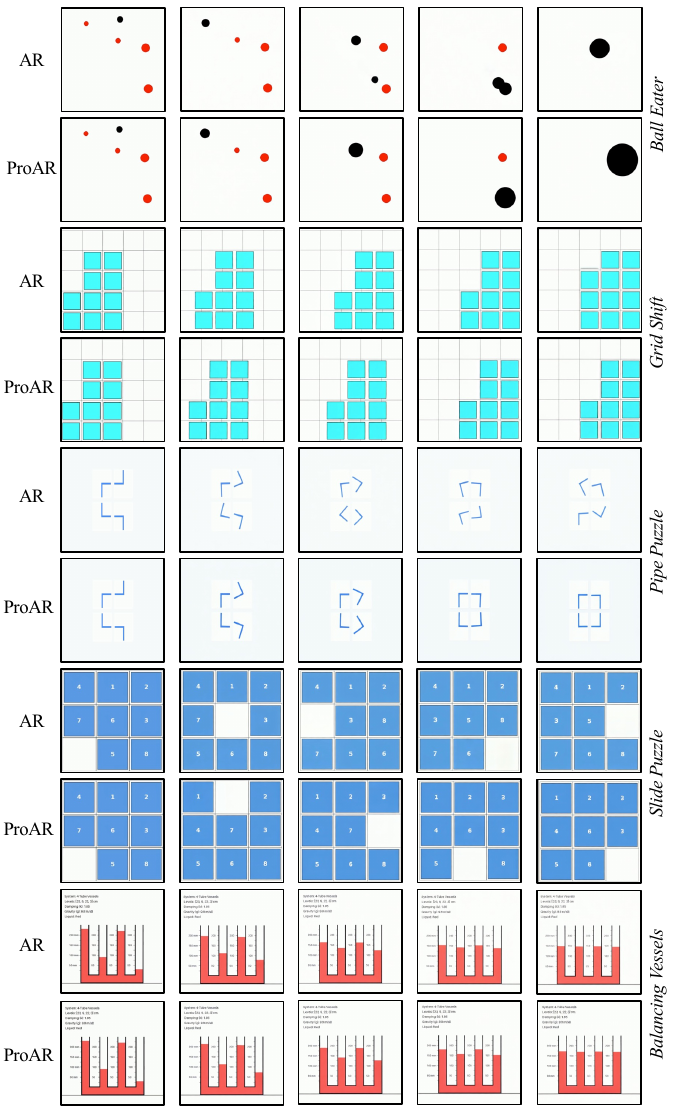}
    \caption{Qualitative Results on VBVR. }
    \label{fig: vbvr2}
\end{figure}

\begin{figure}[!h]
    \centering
    \includegraphics[width=0.9\linewidth]{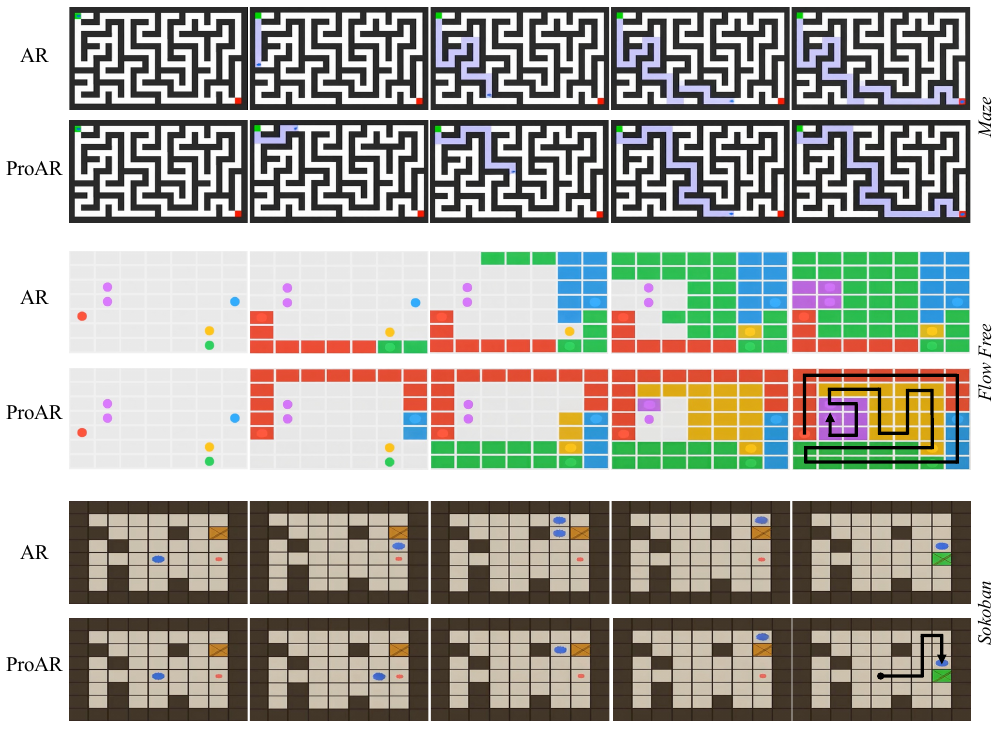}
    \caption{Qualitative Results on VideoRLVR. }
    \label{fig: videorlvr1}
\end{figure}

\begin{figure}[!h]
    \centering
    \includegraphics[width=0.9\linewidth]{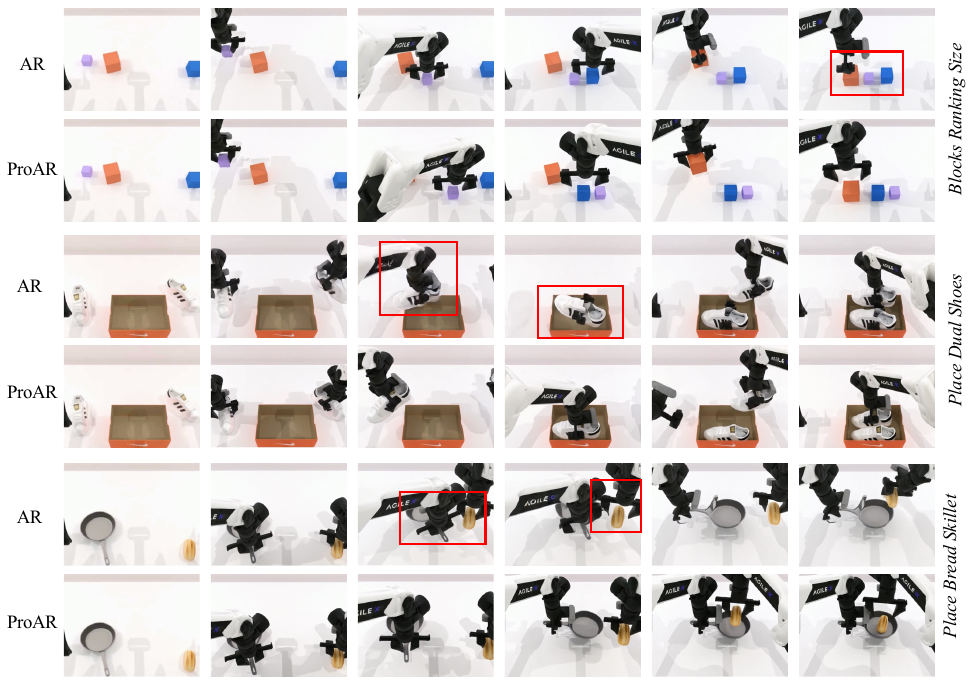}
    \caption{Qualitative Results on WorldArena. }
    \label{fig: worldarena1}
\end{figure}

We provide additional visualization results on VBVR (Fig. \ref{fig: vbvr1} and \ref{fig: vbvr2}), VideoRLVR (Fig. \ref{fig: videorlvr1}), and WorldArena (Fig. \ref{fig: worldarena1})

\end{document}